\documentclass[lettersize,journal]{IEEEtran}
\IEEEoverridecommandlockouts
\pdfoutput=1
\usepackage{cite}
\usepackage{amsmath,amssymb,amsfonts}
\usepackage{algorithmic}
\usepackage[pdftex]{graphicx}
\graphicspath{{../pdf/}{../jpeg/}}
\DeclareGraphicsExtensions{.pdf,.jpeg,.png}
\usepackage{lipsum}
\usepackage{multirow}
\usepackage{textcomp}
\usepackage[table]{xcolor}
\usepackage{tabularx}
\usepackage[ruled]{algorithm2e}
\usepackage{url}
\usepackage{multirow}
\usepackage{longtable} 
\usepackage{booktabs} 
\usepackage{array}
\usepackage[bookmarks=false]{hyperref}
\usepackage{verbatim}
\usepackage{ifpdf}

\begin{document}

\title{Beyond 5G Network Failure Classification for Network Digital Twin Using Graph Neural Network}
\author{
    Abubakar Isah,~\IEEEmembership{Student Member,~IEEE,}
    Ibrahim Aliyu,~\IEEEmembership{Member,~IEEE,}\\
    Jaechan Shim,
    Hoyong Ryu,
    Jinsul Kim,~\IEEEmembership{Member,~IEEE,}\\
    
\thanks{This work was supported in part by the Institute of Information and Communications Technology Planning and Evaluation (IITP) grant funded by the Korean government (MSIT) (Project Name: Development of digital twin-based network failure prevention and operation management automation technology Project Number: RS-2024-00345030, Contribution Rate, 50\%), and in part supported by the Innovative Human Resource Development for Local Intellectualization program through the IITP grant funded by the Korean government (MSIT) (IITP-2024-RS-2022-00156287, 50\%)(Corresponding author: Jinsul Kim).}
\thanks{Abubakar Isah is with HyperIntelligence Media Network Platform Laboratory, Department of Intelligent Electronics and Computer Engineering, Chonnam National University, 61186, Gwangju South Korea, and also with Department of Computer Science, Ahmadu Bello University, Zaria (e-mail: abubakarisah@jnu.ac.kr).}
\thanks{Ibrahim Aliyu and Jinsul Kim are with HyperIntelligence Media Network Platform Laboratory, Department of Intelligent Electronics and Computer Engineering, Chonnam National University, 61186, Gwangju South Korea (e-mail: aliyu@jnu.ac.kr; jsworld@jnu.ac.kr).}
\thanks{Jaechan Shim and Hoyong Ryu are senior researchers at the Electronics and Telecommunications Research Institute (ETRI), South Korea (e-mail: jcshim@etri.re.kr; hyryu@etri.re.kr).}}

\maketitle

\begin{abstract}
Fifth-generation (5G) core networks in network digital twins (NDTs) are complex systems with numerous components, generating considerable data. Analyzing these data can be challenging due to rare failure types, leading to imbalanced classes in multiclass classification. To address this problem, we propose a novel method of integrating a graph Fourier transform (GFT) into a message-passing neural network (MPNN) designed for NDTs. This approach transforms the data into a graph using the GFT to address class imbalance, whereas the MPNN extracts features and models dependencies between network components. This combined approach identifies failure types in real and simulated NDT environments, demonstrating its potential for accurate failure classification in 5G and beyond (B5G) networks. Moreover, the MPNN is adept at learning complex local structures among neighbors in an end-to-end setting. Extensive experiments have demonstrated that the proposed approach can identify failure types in three multiclass domain datasets at multiple failure points in real networks and NDT environments. The results demonstrate that the proposed GFT-MPNN can accurately classify network failures in B5G networks, especially when employed within NDTs to detect failure types.
\end{abstract}


\begin{IEEEkeywords}
\textbf{Beyond fifth generation (B5G), class imbalance, graph Fourier transform (GFT), message-passing neural network (MPNN), network digital twin (NDT)}
\end{IEEEkeywords}

\IEEEpeerreviewmaketitle

\section{Introduction}

\IEEEPARstart{T}{he} emergence of fifth-generation (5G) networks, and robust core network systems are paramount for the evolution to Industry 5.0. According to recent studies, 5G connections are projected to increase over 100-fold, from approximately 13 million in 2018 to 1.4 billion by 2023 \cite{Cisco:2020}. This growth is expected to continue with sixth-generation (6G) networks and the advent of the Industry 5.0 era, characterized by the vision of the Internet of Everything. In this context, network failure classification for 5G and beyond is critical. As the backbone of communication infrastructure, reliable and resilient networks are essential for seamless operation in diverse applications in environments ranging from daily life to industrial settings. The core network management center can identify and classify network failures by applying deep learning algorithms and advanced analytics, enabling preemptive measures to ensure uninterrupted service delivery \cite{wang2022spatial}.

The digital twin (DT) is a powerful technology that has emerged as a promising solution for connecting physical spaces with digital systems \cite{grieves2014digital, isah2023digital}. This technology involves creating digital replicas of physical entities, such as devices, machines, and objects, using historical data and real-time operational information. In addition, this technology facilitates close monitoring, real-time interaction, and dependable communication between the digital and physical worlds. Research on DTs \cite{yu2023digital} has garnered significant attention, with numerous researchers exploring its applications in such fields as aviation, 6G networks, intelligent manufacturing, and data generation \cite{wu2021digital, isah2023towards}. Additionally, DT has been recognized as one of the 12 representative use cases for future networks by the  International Telecommunication Union Network 2030 focus group \cite{cases2020key}. However, setting up a replica of a network event to mimic every potential pattern in network DTs (NDTs) is challenging.

Many real-world data, including those from social networks, the Internet of Things (IoT), and finances, are represented as graphs; hence, graph neural networks (GNNs) have gained popularity in many real-world applications \cite{smilkov2017smoothgrad}, \cite{zhou2016learning}, and \cite{selvaraju2017grad}. Numerous tasks related to graphs have been extensively researched, including node classification \cite{dabkowski2017real}, graph classification \cite{yuan2020interpreting}, and link predictions \cite{olah2017feature}. Moreover, a variety of advanced GNN procedures, such as graph pooling \cite{olah2018building}, \cite{yang2019xfake}, \cite{du2018towards}, graph convolution \cite{wang2020icapsnets}, \cite{du2019attribution}, and graph attention \cite{yuan2019interpreting}, \cite{du2019techniques}, have been suggested to enhance the performance of GNNs.
Nevertheless, there is still lacking on graph models in compared to communication network domains \cite{isah2024graph}, which is crucial for comprehending deep GNNs. Several methods, such as RouteNet \cite{ferriol2023routenet} and RouteNet-Fermi \cite{galmes2022routenet}, have been proposed to model the NDT for network performance analysis. The benefit of data-driven models \cite{isah2023datadriven} is their ability to accurately represent a wide range of intricate network properties with previously unheard-of accuracy as they are trained on real-world data \cite{zhang2024knowledge}.

The class imbalance problem in traditional feature-based supervised learning settings has recently been the focus of research efforts. These efforts have primarily explored oversampling the minority class \cite{liu2020alleviating} or undersampling the majority class \cite{peng2019trainable}, which are resampling techniques that balance the number of instances. Although class-imbalanced supervised learning in traditional feature space has been extensively researched, scarce research on GNN techniques has focused solely on this problem. A recent attempt in \cite{huang2022graph} has been made to address class imbalance on graph data.

This paper proposes a GNN-based imbalanced learning approach for 5G network failure classification that employs the graph Fourier transform (GFT) to address these problems. The 5G network is modeled as a graph. The GFT, a core component of the message-passing neural network (MPNN) architecture, iteratively updates node features by considering the intrinsic node properties and features of its connected neighbors. This iterative process allows information to propagate across the network, mimicking a form of communication that captures the inherent relationships within the network structure. The proposed model achieves superior accuracy in pinpointing both the type and location of failures within the 5G network by exploiting these learned representations.

\subsection{Related work}
Deep learning-driven failure predictions have been garnering interest due to their effectiveness in managing massive volumes of data generated by 5G networks and their ability to understand the complex structural characteristics of these networks. This section presents two areas of research: one focusing on learning-based 5G failure prediction, and imbalance learning techniques, and the other investigating GFT to capture the complex structure of NDTs.

\subsubsection{Learning-based failure prediction}
   The primary benefit of applying deep learning to network modeling \cite{habibi2019comprehensive} is its data-driven approach, which may enable it to simulate the complex nature of the actual networks. Most of the of the current suggestions in \cite{islam2022deep} and \cite{said2020network} employ conventional fully connected neural networks. Nevertheless, the primary constraint of these methods is their non-generalizability to alternative network topologies and configurations, such as routing. In this context, more recent studies have suggested more complex neural network models, such as GNNs \cite{shojaee2020safeguard, soltanzadeh2021rcsmote}, convolutional neural networks \cite{said2020network}, and variational autoencoders \cite{ding2018opportunities}. However, these models have different objectives and ignore the essential elements of actual networks from the DT perspective. The concept of broadly applying neural networks to graphs has gained attention. Convolutional networks were extended to graphs in the spectral domain \cite{bruna2013spectral}, where filters are applied to frequency modes in a graph, which are determined using the GFT. The eigenvector matrix of a graph Laplacian must be multiplied for this transformation. One study \cite{defferrard2016convolutional} parameterized the spectrum filters as Chebyshev polynomials of eigenvalues, resulting in efficient and localized filters and reducing the computational burden. The mentioned spectral formulations are disadvantageous in that they only apply to graphs with a single structure since they depend on the constant spectrum of the graph Laplacian. Conversely, spatial formulations are not limited to a specific graph topology. Generalizing neural networks to graphs is studied in two ways: a) given a single graph structure, signals in graph forms or labels of individual nodes \cite{scarselli2008graph, bruna2013spectral, lei2017deriving, li2015gated, defferrard2016convolutional, kipf2016semi} b) given a set of graphs with different structure and sizes, the preditions of the class labels of unseen graphs are learned \cite{duvenaud2015convolutional, atwood2016diffusion, niepert2016learning}.
\subsubsection{Imbalance learning technique}
Data-level and algorithm-level methods are the two main categories of class-imbalanced learning techniques. Before building classifiers, data-level approaches preprocess training data to reduce inequality \cite{li2021novel}. These strategies include undersampling majority classes and oversampling minority classes. To address the issue of class imbalance, algorithm-level approaches alter the model's fundamental learning or decision-making process. Algorithm-level techniques can be broadly classified into three categories: threshold moving, cost-sensitive learning, and new loss functions \cite{johnson2019survey}. Many approaches have been proposed at the data level. SMOTE \cite{chawla2002smote} is one such technique that creates fake minority samples in the feature space by interpolating current minority samples and their closest minority neighbors. SMOTE's primary flaw is that it creates new synthetic samples without taking neighborhood samples of the majority classes into account, which can lead to increased class overlapping and extra noise \cite{koziarski2019radial}. Based on the SMOTE principle, several variations have been put forth, such as Borderline-SMOTE \cite{han2005borderline} and Safe-Level-SMOTE \cite{bunkhumpornpat2009safe}, which enhance the original approach by taking majority class neighbors into account. Safe-Level-SMOTE creates safe zones to prevent oversampling in overlapping or noisy regions, whereas Borderline-SMOTE only samples the minority samples close to the class borders. To address class-imbalanced challenges, \cite{gilmer2017neural} modified the parameters of the model learning process that favored classes with fewer data. For example, minority samples can contribute more to the loss thanks to the development of Focal Loss in \cite{lin2017focal}. Deep networks were trained on imbalanced datasets using a novel loss function termed Mean Squared False Error (MSFE), which was proposed in \cite{wang2016training}. 
In order to particularly handle the class-imbalanced problem on graph data, two new models have been developed recently: Dual-Regularized Graph Convolutional Network (DRGCN) \cite{shi2020multi} and GraphSMOTE \cite{zhao2021graphsmote}. Adversarial training and distribution alignment are the two regularization types that DR-GCN uses, whereas GraphSMOTE uses SMOTE \cite{chawla2002smote} to create synthetic nodes for every minority class and link prediction-based techniques \cite{tan2015online} to create new edges.

\subsubsection{Graph Fourier transform}
This work extracts features by transforming the dataset using GFT to identify the global vertex labels for the whole graph and a local variable using the MPNN for each level while training the model. Although achieving a competitive result by using multiclass classification, the drawback of this approach includes heavy data preprocessing and reliance on creating an NDT in 5G settings.
This pattern is exemplified by the ITU dataset employed in this research \cite{Albert:2023}. The KDDI network of the DT datasets exhibits a significant class imbalance, with only 0.2\% of the 16 failure types representing failures, whereas the normal case comprises 67\% of the data. Consequently, when applied to such imbalanced datasets without addressing the data imbalance, deep learning models tend to favor the majority class \cite{johnson2019survey}. GFT-MPNN is also related to GNNs \cite{scarselli2008graph, li2015gated} and Neural Graph Fingerprints \cite{duvenaud2015convolutional} regarding how to extract node features. However, GNNs supervise a single node to perform graph-level classification, and neural graph fingerprints use summed node features \cite{smilkov2017smoothgrad}. In comparison, GFT transforms the data and efficiently captures the global network structure and dependencies between nodes, whereas MPNN offers more flexibility in learning localized relationships in multiclass settings.

\subsection{Motivation and Contributions}

The B5G and future networks are incredibly complex, making efficiently diagnosing and addressing failures challenging. Applying GNNs allows for an effective analysis of the structural information in network data to classify network failures accurately. By integrating these technologies with DT technology, operators can create virtual replicas of network systems, enabling real-time monitoring, analysis, and simulation to identify and mitigate failure modes. This comprehensive approach aims to meet the strict requirements of B5G networks while laying the groundwork for advancements in future telecommunications technologies. 

The contributions of this work and the proposed GFT-MPNN include the following:

\begin{enumerate}
    \item The proposed approach, GFT-MPNN, integrates the GFT with an MPNN model to address the challenge of imbalanced data across different domains, ensuring robust performance in classifying network failures in real-world and NDT data.
    
    \item The MPNN is used as a feature extraction method to model dependencies between network structure nodes and is adept at learning the complex local structures between neighbors in an end-to-end setting for multiclass failure classification tasks.
    
    \item We conduct a comprehensive end-to-end evaluation of the GFT-MPNN framework using real and simulated datasets. This evaluation encompasses training and testing phases, assessing the model's performance in practical scenarios.

\end{enumerate}
The paper was structured as follows: Section II explains the problem statement and the details regarding the challenge by ITU. Next, Section III provides the system architecture and strengths of the proposed approach. Then, Section IV details the datasets and results. Finally, Section V presents the conclusion.


\section {Problem Formulation and GFT-MPNN Framework}

\subsection{Problem Statement}

Due to the increased number of components in the networks, which increases the operational cost, ensuring stable and high-quality internet connectivity has become increasingly critical for network operators as 5G networks continue to expand. However, machine learning presents a promising approach to address unexpected failures in the IP core network, offering automation and cost-effectiveness. Based on the ITU Challenge "ITU-ML5G-PS-008: Network failure classification model using the digital twin network" \cite{Junichi:2023}, this research aims to employ GNNs designed to analyze networks from end-to-end and other interconnected systems to classify network failure types and failure points more accurately than typical machine learning models.

We formulated the network failure injection as a classification problem, focusing on learning meaningful embeddings from the generated data to increase detection accuracy and identify different failure types. In the network structure, we define a graph $G=(V,E)$ with an adjacent matrix $A = \{e_{vu} | \forall v, u \in V\} \in \mathbb{R}^{N \times N}$, and $N$ representing the number of nodes, letting $G = (V, E)$ represent an input graph with a set of nodes $V$. If $e_{vu} = 1$, then $(v, u) \in E$; otherwise, $(v, u) \in E'$. We assume the set of feature vectors associated with each network component $v$ are $D$-dimensional represented by $F = \{f_v | v \in V\} \in \mathbb{R}^D$.

We considered the failure points as nodes in the network graph for network failure classification. Each node is associated with various features, such as performance metrics or operational data, reflecting its behavior within the network. The edges connecting pairs of nodes represent the dependencies and relationships between them.

\begin{figure}
    \centering
    \includegraphics[width=1.0\linewidth]{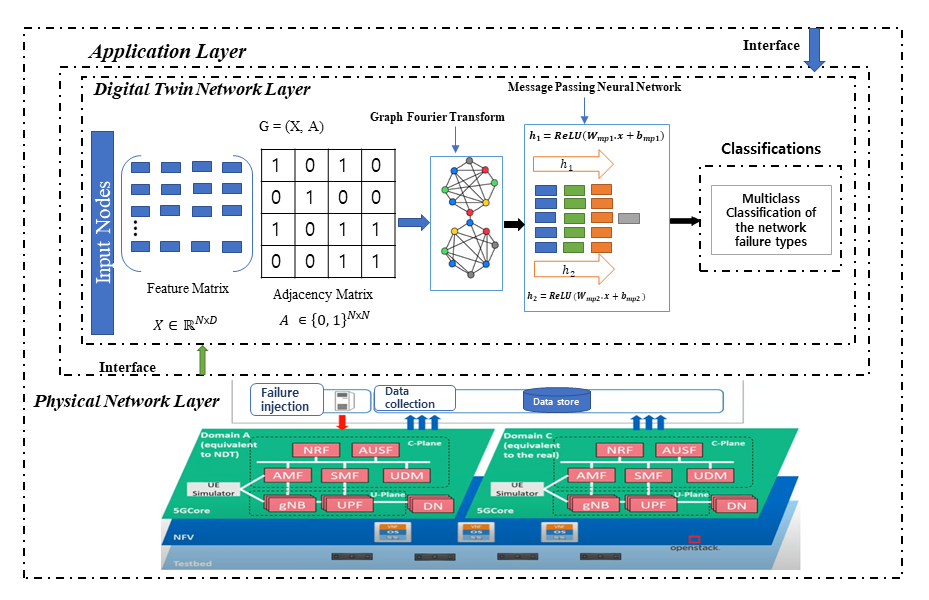}
    \caption{Beyond fifth-generation network failure classification using a network of digital twin systems}
    \label{fig: DTN 5G}
\end{figure}

\subsection{Proposed Model}
Fig. \ref{fig: DTN 5G} illustrates the general architecture of the proposed model. We applied GNNs to determine edge memberships within different connection types (i.e., oper-status, phys-address e.t.c). The provided NDT data were input into a neural network within the graph module. The GFT captures global network properties and smooths noisy features, whereas the MPNN extracts fine-grained information regarding node interactions and local graph structures. The graph module then processes these inferred connections to create a multi-layer graph structure. The feature extraction module employs Fourier transform techniques to transform the 5G network data into meaningful features. Finally, the GFT-MPNN combines these extracted features and the learned graph structure to deliver the final multi-class classification outcome. The following sections describe the concepts of the GNN and its variants, consisting of the GFT and MPNN, which are critical components of the network failure classification problem.

\begin{table*}[h]
\centering
\caption{Explanation of Variables}
\label{tab:variables_explanation}
\begin{tabular}{|l|p{5.5cm}|l|p{5.5cm}|}
\hline
\textbf{Variable} & \textbf{Explanation} & \textbf{Variable} & \textbf{Explanation} \\ 
\hline
$G = (V,E)$ & Network digital twin system graph. & $N$ & Total number of vertices. \\
$V_n$ & Vertex set of graph $G_n$, indexing nodes. & $y_{ij}$ & $j$-th component of one-hot label for $v_i$. \\
$N_n$ & Number of vertices in $G_n$. & $f_{ij}$ & Predicted probability of $v_i$ in $j$-th class. \\
$L_n$ & Laplacian matrix of $G_n$ capturing structure and connectivity. & $W_{fc1}, b_{fc1}, W_{fc2}, b_{fc2}$ & GFT model linear layer parameters. \\
$\lambda^{(n)}_k$ & $k$-th eigenvalue of $\mathcal{L}^{(n)}$. & $W_{mp1}, b_{mp1}, W_{mp2}, b_{mp2}$ & MPNN message passing layer parameters. \\
$\phi^{(n)}_k$ & Orthonormal eigenfunctions of the Laplacian. & $h_i$ & Hidden representation at layer $i$. \\
$\mathcal{G}_1 \times \mathcal{G}_2$ & Product graph, features associated with nodes. & $\hat{m}, \hat{v}$ & Moment estimates in Adam optimizer. \\
$\sigma(L_1 \oplus L_2)$ & Eigenvalues of $L_1 \oplus L_2$. & $\eta$ & Learning rate. \\
$(i_1, i_2)$ & Vertex indices in $\mathcal{G}_1$ and $\mathcal{G}_2$. & $y$ & Output vector of size $(N, C)$, $C$ is number of classes. \\
$(k_1, k_2)$ & Indices for eigenvalues and eigenfunctions. & & \\
$\theta$ & GNN model parameters to be learned. & & \\
\hline
\multicolumn{4}{|p{17.5cm}|}{$f^{(l)}$ is the aggregation function at layer $l$, $N(v_i)$ represents the neighbors of $v_i$, and $h_i^{(0)} = x_i$ is the initial feature vector.} \\
\hline
\end{tabular}
\end{table*}

\subsection{ Graph Neural Network}
The GNN can transform the graph structure data into standard representations and input them into neural networks for training. This approach allows the GNN to propagate the information of nodes and edges efficiently to neighboring nodes or even the entire graph. With the GNN, a node representation vector is calculated by iteratively aggregating and converting the representation vectors of its nearby nodes. The GNN uses the neighbor aggregation approach in \cite{lessan2019hybrid}.
The network infrastructure is represented as a graph $G = (V,E)$, where $V$ represents the set of nodes (network components) and $E$ represents the set of edges (connections). Each node $v_i$ in the graph is associated with a feature vector xi, which captures relevant information about the corresponding network component. The graph can be represented using an adjacency matrix $A$, where $A_{ij} = 1$ if a connection exists between nodes $v_i$ and $v_j$, and $A_{ij} = 0$ otherwise. Node features are represented by the feature matrix $X$, where $X$ has the dimensions $|V| \times D$, with $|V|$ represents the number of nodes, and $D$ denotes the dimensionality of the feature vectors $X = [x_1, x_2, \ldots, x_{|V|}]$. 

Many network components depend upon one another due to their interconnectivity \cite{ferriol2023routenet}.
The GNN architecture consists of multiple layers, each processing information from neighboring vertices to extract hierarchical representations. We let $h_i^{(l)}$ denote the representation of vertex $v_i$ at layer $l$ of the GNN. Information is aggregated from the neighboring vertices and transformed using learnable parameters. The update equation for $h_i^{(l)}$ can be expressed as follow:

\begin{equation}
h_i^{(l)} = f^{(l)} \left( h_i^{(l-1)}, \{ h_j^{(l-1)} \}_{j \in N(v_i)} \right),
\end{equation}

One significant challenge is preparing NDT datasets that can consistently replicate real-world failures. This difficulty arises from several factors, including limited data on real-world systems and the failure mode complexity.

\section{MODELING THE PROPOSED GFT-MPNN}

As depicted in Fig. \ref{fig: DTN 5G}, the system architecture of the proposed GFT-MPNN consists of three main streams: 1) data preprocessing, 2) the strengths of the GFT-MPNN model, and 3) model training algorithm. The proposed method uses end-to-end generalization. This section summarizes the proposed method, outlining each module and its functionality.

\subsection{Graph Fourier Transform}
We considered an undirected weighted graph \( G = (V, E, w) \), where \( V = \{0, 1, \ldots, N-1\} \) is the vertex set, \( E \) is the edge set, and \( w(i,j) \) is the weight function satisfying \( w(i,j) = 0 \) for any \( (i,j) \) not in \( E \). We assume all graphs are simple, meaning they have no loops or multiple edges.

Three matrices associated with \( G \) are important: the adjacency matrix \( W \), a degree matrix \( D \), and a Laplacian matrix \( L = D - W \). The Laplacian matrix is essential for Graph Fourier Transforms (GFTs).

The Laplacian matrix \( L \) is real, symmetric, and positive-semidefinite, thus, it possesses nonnegative eigenvalues \( \lambda_0, \ldots, \lambda_{N-1} \) and the corresponding orthonormal eigenfunctions \( u_0, \ldots, u_{N-1} \). These eigenfunctions satisfy the following equation:

\begin{equation}
L \begin{bmatrix}
u_k(0) \\
\vdots \\
u_k(N-1)
\end{bmatrix} = \lambda_k \begin{bmatrix}
u_k(0) \\
\vdots \\
u_k(N-1)
\end{bmatrix}
\end{equation}

This study assumes that the eigenvalues are sorted in ascending order (\( \lambda_0 \leq \ldots \leq \lambda_{N-1} \)). Notably, \( \lambda_0 \) is strictly zero because the rowwise sums of \( L \) are all zero. The spectrum of the matrix \( L \) is denoted by \( \sigma(L) \).

For \( k = 0, \ldots, N-1 \), orthonormality implies that the sum of products of corresponding eigenfunctions is equal to the Kronecker delta function \(\delta(i,j)\), where \(\delta(i,j)\) equals 1 if \(i = j\) and is zero otherwise. This study assumes that eigenvalues are arranged in ascending order, such that \(\lambda_0 \leq \ldots \leq \lambda_{N-1}\). In addition, \(\lambda_0\) is strictly zero because the sum of rows in \(L\) equals zero. The spectrum of the matrix, denoted as \(\{ \lambda_k \}_{k=0}^{N-1}\), is represented as \(\sigma(L)\).
The GFT in \cite{kurokawa2017multi} of a graph feature \( f : V \rightarrow \mathbb{R} \) is defined as \( \hat{f} : \sigma(L) \rightarrow \mathbb{C} \), where \( \sigma(L) \) represents the spectrum of the graph Laplacian \( L \). It is expressed as follows:

\begin{equation}
    \hat{f}(\lambda_k) = \langle f, u_k \rangle = \sum_{i=0}^{N-1} f(i) u_k(i),
\end{equation}

for \( k = 0, \ldots, N-1 \). where, \( u_k \) denotes the orthonormal eigenfunctions of \( L \), and the GFT represents a feature expansion using these eigenfunctions. The inverse GFT is given by,

\begin{equation}
    f(i) = \sum_{k=0}^{N-1} \hat{f}(\lambda_k) u_k(i),
\end{equation}

On cycle graphs, the GFT is equivalent to the discrete Fourier transform (DFT).

When a graph Laplacian has non-distinct eigenvalues, the functions generated by the GFT may not be defined well, resulting in multi-valued functions. For instance, if two orthonormal eigenfunctions, \( u \) and \( u_0 \) correspond to the same eigenvalue \( \lambda \), then the spectral component \( \hat{f}(\lambda) \) can have two distinct values: \( \langle f,u \rangle \) and \( \langle f,u_0 \rangle \).

\subsection{Twin Graph Fourier Transform}
The Cartesian product \( G_1 \square G_2 \) of graphs \( G_1 = (V_1, E_1, w_1) \) and \( G_2 = (V_2, E_2, w_2) \) is a graph with the vertex set \( V_1 \times V_2 \) and the edge set \( E \) defined as follows. For any \((i_1, i_2)\) and \((j_1, j_2)\) in the vertex set \( V_1 \times V_2 \), these are connected by an edge if either \((i_1, j_1)\) is in \( E_1 \) and \(i_2 = j_2\) or \(i_1 = j_1\) and \((i_2, j_2)\) is in \( E_2 \). The weight function \( w \) is defined as follows:

\begin{equation}
    w((i_1, i_2), (j_1, j_2)) = w_1(i_1, j_1) \delta(i_2, j_2) + \delta(i_1, j_1) w_2(i_2, j_2)
\end{equation}

where, \( \delta \) denotes the Kronecker delta function. The graphs \( G_1 \) and \( G_2 \) are referred to as the factor graphs of \( G_1 \square G_2 \).

The adjacency, degree, and Laplacian matrices of a Cartesian product graph can be derived from those of its factor graphs. Two factor graphs, \( G_1 \) and \( G_2 \), with vertex sets \( V_1 = \{0, 1, \ldots, N_1 - 1\} \) and \( V_2 = \{0, 1, \ldots, N_2 - 1\} \), respectively, are considered, each with adjacency matrix \( W_1 \) and \( W_2 \), degree matrix \( D_1 \) and \( D_2 \), and Laplacian matrix \( L_1 \) and \( L_2 \). When the vertices of the Cartesian product graph are ordered lexicographically, such as \((0, 0), (0, 1), (0, 2), \ldots, (N_1 - 1, N_2 - 1)\), the adjacency, degree, and Laplacian matrices of \( G_1 \square G_2 \) can be expressed as \( W_1 \oplus W_2 \), \( D_1 \oplus D_2 \), and \( L_1 \oplus L_2 \), respectively, where operator \( \oplus \) denotes the Kronecker sum.

\textbf{Definition 1:} The Kronecker sum is defined by \( A \oplus B = A \otimes I_n + I_m \otimes B \) for matrices \( A \in \mathbb{R}^{m \times m} \) and \( B \in \mathbb{R}^{n \times n} \), where \( I_n \) represents the identity matrix of size \( n \).

The Kronecker sum has a valuable characteristic that decomposes an eigenproblem involving the Laplacian matrix of a product graph into eigenproblems of Laplacian matrices of the factor graphs. We assumed that the Laplacian matrix \( L_n \) of each factor graph has nonnegative eigenvalues \(\{\lambda^{(n)}_k\}_{k=0}^{N_n-1}\) and orthonormal eigenfunctions \(\{u^{(n)}_k\}_{k=0}^{N_n-1}\) for \(n=1,2\). In this case, the Kronecker sum \( L_1 \oplus L_2 \) yields an eigenvalue of \(\lambda^{(1)}_k + \lambda^{(2)}_k\) and the corresponding eigenfunction \(u^{(1)}_k \otimes u^{(2)}_k : V_1 \times V_2 \rightarrow \mathbb{C}\), where \(\oplus\) denotes the Kronecker sum. This eigenfunction satisfies the following:

\begin{align}
& (L_1 \oplus L_2) \left[ \begin{array}{c} u^{(1)}_k(0) u^{(2)}_k(0) \\ u^{(1)}_k(0) u^{(2)}_k(1) \\ \vdots \\ u^{(1)}_k(N_1-1) u^{(2)}_k(N_2-1) \end{array} \right] \\
& = (\lambda^{(1)}_k + \lambda^{(2)}_k) \left[ \begin{array}{c} u^{(1)}_k(0) u^{(2)}_k(0) \\ u^{(1)}_k(0) u^{(2)}_k(1) \\ \vdots \\ u^{(1)}_k(N_1-1) u^{(2)}_k(N_2-1) \end{array} \right]
\end{align}

This property holds for any \(k_1 = 0, \ldots, N_1-1\) and \(k_2 = 0, \ldots, N_2-1\). The resulting eigenvalues \(\{\lambda^{(1)}_k + \lambda^{(2)}_k\}\) are nonnegative, and the corresponding eigenfunctions \(\{u^{(1)}_k \otimes u^{(2)}_k\}\) are orthonormal. These conclusions can be straightforwardly derived from the fundamental properties of the Kronecker product.

A similar approach can be applied to decompose an eigenproblem related to an adjacency matrix of a product graph. Assuming that the adjacency matrix $W_n$ has eigenvalues $\{\mu^{(n)}_k\}_{k=0}^{N_n-1}$ and orthonormal eigenfunctions $\{\mathbf{v}^{(n)}_k\}_{k=0}^{N_n-1}$ for $n=1,2$, the Kronecker sum $W_1 \oplus W_2$ yields the eigenvalues $\{\mu^{(1)}_k + \mu^{(2)}_l\}$ and corresponding eigenfunctions $\{\mathbf{v}^{(1)}_k \otimes \mathbf{v}^{(2)}_l\}$ defined on $V_1 \times V_2$.

The Cartesian product graph is represented as $G_1 \times G_2$, where $G_n$ for $n=1,2$ is an undirected weighted graph with a vertex set $V_n = \{0,1,\dots,N_n-1\}$, and assuming its graph Laplacian $L_n$ has ascending eigenvalues $\{\lambda_k^{(n)}\}_{k=0}^{N_n-1}$ and corresponding orthonormal eigenfunctions $\{u_k^{(n)}\}_{k=0}^{N_n-1}$. After talking about product graphs. The GFT of the graph feature $f:V_1 \times V_2 \rightarrow \mathbb{R}$ on the product graph $G_1 \times G_2$ is represented by a spectrum $\hat{f}:\sigma(L_1 \oplus L_2) \rightarrow \mathbb{C}$, provided by:

\begin{equation}
    \hat{f} : (\sigma(L_1) \otimes \sigma(L_2)) \rightarrow \mathbb{C}
\end{equation}

\begin{equation}
\hat{f}(\lambda^{(1)}_{k_1} + \lambda^{(2)}_{k_2}) = \sum_{i_1=0}^{N_1-1} \sum_{i_2=0}^{N_2-1} f(i_1,i_2) \cdot u^{(1)}_{k_1}(i_1) \cdot u^{(2)}_{k_2}(i_2)
\end{equation}

for $k_1=0,\dots,N_1-1$ and $k_2=0,\dots,N_2-1$, and its inverse is:

\begin{equation}
f(i_1,i_2) = \sum_{k_1=0}^{N_1-1} \sum_{k_2=0}^{N_2-1} \hat{f}(\lambda^{(1)}_{k_1} + \lambda^{(2)}_{k_2}) \cdot u^{(1)}_{k_1}(i_1) \cdot u^{(2)}_{k_2}(i_2)
\end{equation}

for $i_1=0,\dots,N_1-1$ and $i_2=0,\dots,N_2-1$. The feature spectrum is not defined as a univariate function on $\sigma(L_1 \oplus L_2)$, but rather as a bivariate function on $\sigma(L_1) \times \sigma(L_2)$. Hence, we interpreted the features on a Cartesian product graph as "2D features" and proposed a twin GFT, yielding a multiclass failure extraction classification in NDT settings.

\textbf{Definition 1} The Twin Graph Fourier Transform (Twin GFT) of a feature $f: V_1\times V_2\rightarrow \mathbb{R}$ on a Cartesian product graph $G_1 \square G_2$ is a spectrum $\hat{f}:\sigma(L_1)\times \sigma(L_2) \rightarrow \mathbb{C}$ defined by the following:

\begin{equation}
    \hat{f}(\lambda_{11},\lambda_{22}) = \sum_{i_1=0}^{N_1-1} \sum_{i_2=0}^{N_2-1} f(i_1,i_2) \cdot u_{k_{11}}^{(1)}(i_1) \cdot u_{k_{22}}^{(2)}(i_2)
\end{equation}

for $k_{11}=0,\dots,N_1-1$ and $k_{22}=0,\dots,N_2-1$, and its inverse is given by:

\begin{equation}
    f(i_1,i_2) = \sum_{k_1=0}^{N_1-1} \sum_{k_2=0}^{N_2-1} \hat{f}(\lambda_{k_{11}}^{(1)} + \lambda_{k_{22}}^{(2)}) \cdot u_{k_{11}}^{(1)}(i_1) \cdot u_{k_{22}}^{(2)}(i_2)
\end{equation}

for $i_1=0,\dots,N_1-1$ and $i_2=0,\dots,N_2-1$.

The twin-GFT is represented as a series of matrix-matrix multiplications. Using $N_1 \times N_2$ matrices $F=(f(i_1,i_2))_{i_1,i_2}$ and $\hat{F}=(\hat{f}(\lambda_{k_{11}}^{(1)} + \lambda_{k_{22}}^{(2)}))_{k_{11},k_{22}}$, the twin-GFT applied to the features $f$ is expressed as follows:

\begin{equation}
\hat{F} = U_1^* FU_2,
\end{equation}

where $U_n$ denotes an $N_n \times N_n$ unitary matrix with $(i,k)$-th element $u_{k}^{(n)}(i)$ for $n=1,2$. Then its inverse is given by:

\begin{equation}
    F = U_1 \hat{F}^2 U_2^*,
\end{equation}

The twin GFT has connections to existing transformations. First, when both factor graphs are cycle graphs, the twin GFT can be equivalent to the 2D DFT and GFT (2D GFT). Second, when a factor graph is a cycle graph, some cycles might be nested within other cycles, creating a complex network structure. The twin GFT is known as a joint graph and temporal Fourier transform which generalizes these existing transformations. 

The suggested twin GFT offers the following benefits over the traditional GFT for a feature on product graphs.

\begin{itemize}
\item \textbf{Multiclass classification resolution}:
The twin GFT addresses the multiple values in the ordinary GFT when dealing with nondistinct eigenvalues. It ensures well-defined spectra even when features are equal but are from different pairs, thus providing clarity in spectrum assignments.

\item \textbf{Reduced computational complexity}:
The twin GFT offers reduced computational complexity on product graphs compared to the conventional GFT. Its time complexity for both the transform and its inverse is $O(N_1^2 N_2 + N_1 N_2^2)$, significantly less than the $O(N_1^2 N_2^2)$ complexity of conventional GFTs. This reduction is attributed to the efficient matrix multiplication algorithms employed in the twin GFT computations.

\item \textbf{Insightful feature analysis}:
The twin GFT enables an insightful analysis of graph features along each factor graph. Its counterpart with the 2-D Fourier transform allows for understanding feature variations across dimensions, aiding in interpreting complex data structures and failure classes.
\end{itemize}

\subsection{GFT-MPNN Training Procedure}

A crucial advantage of the GNNs for the NDT failure classification is the adaptive architecture. In this context, the graph represents the NDT with nodes representing specific network elements and edges representing their connections. The hidden state, $h_i$, of each node initially captures features relevant to failure prediction, such as bridge down, interface down, or packet loss. Through message passing, nodes iteratively update their hidden states by aggregating information from their connected neighbors. This process permits failures and their effects to propagate across the network, mirroring real-world failure scenarios. As the number of iterations increases, information can potentially traverse the entire network, capturing complex dependencies even between distant nodes in the DT. This ability to apply a graph structure is crucial for the multiclass classification task, because it allows the GNN to differentiate between various failure types based on the propagation patterns observed within the network.

The input feature vectors $\mathbf{x}$ for each node are transformed linearly using weight matrix $\mathbf{W}_{\text{mp1}}$ and bias vector $\mathbf{b}_{\text{mp1}}$. This transformation is denoted as $\mathbf{h}_1 = \text{ReLU}(\mathbf{W}_{\text{mp1}} \cdot \mathbf{x} + \mathbf{b}_{\text{mp1}})$.

The output of the linear transformation is then passed through the Rectified Linear Unit (ReLU) activation function, denoted as $\text{ReLU}(\cdot)$, which introduces nonlinearity into the model. The ReLU activation function is applied element-wise to the result of the linear transformation. The output of this operation represents the hidden states $\mathbf{h}_1$ for all nodes in the graph, capturing the extracted features and aggregated information from neighboring nodes.

The first message-passing layer performs the following operation:

\begin{equation}
    \mathbf{h}_1 = \text{ReLU}(\mathbf{W}_{\text{mp1}} \cdot \mathbf{x} + \mathbf{b}_{\text{mp1}}),
\end{equation}

where $\mathbf{h}^{(1)}$ is the hidden state vector obtained after the first message-passing layer, $\mathbf{x}$ denotes the input feature vector for each node, $\mathbf{W}_{\text{mp1}}$ is the weight matrix for the linear transformation, and $\mathbf{b}_{\text{mp1}}$ indicates the bias vector, and $\text{ReLU}(\cdot)$ is the Rectified Linear Unit activation function. Algorithm \ref{alg:two} represents the GFT-MPNN training process for the NDT in the multiclass classification of failure types. 

\begin{algorithm}[hbt!]
\caption{Proposed GFT-GNN Model}
\label{alg:two}

\textbf{Input:}
$X$, $y_{\text{true}}$, $A$, $W_{\text{mp1}}$, $b_{\text{mp1}}$, $W_{\text{mp2}}$, $b_{\text{mp2}}$, GFT, $\eta$, $\text{num\_epochs}$

\textbf{Output:}
$y_{\text{pred}}$

\textbf{Initialization:}
\begin{itemize}
    \item Input feature vector: $X$ (shape: $N \times D$)
    \item True labels: $y_{\text{true}}$ (shape: $N$)
    \item Adjacency matrix: $A$ (shape: $N \times N$)
    \item Learning rate: $\eta$
    \item Number of epochs: $\text{num\_epochs}$
\end{itemize}

\textbf{Training loop:}
\For{$\text{epoch}$ in range$(\text{no.\_epochs})$}{
    \textbf{forward pass:}
    \begin{enumerate}
        \item Perform Min-Max scaling on $X$: $X_{\text{norm}} = \text{MinMaxScaler}(X)$
        \item Convert data to float32: $X_{\text{float}} = \text{float32}(X_{\text{norm}})$, $y_{\text{int}} = \text{int64}(y_{\text{true}})$
        \item First message passing layer: $h_1 = \text{ReLU}(W_{\text{mp1}} \cdot X_{\text{float}} + b_{\text{mp1}})$
        \item Message passing: $h_2 = \text{ReLU}(A \cdot h_1)$
        \item Graph Fourier transform: $y_{\text{pred}} = \text{GFT}(h_2, A)$
    \end{enumerate}
    \textbf{Loss calculation:}
    \begin{enumerate}
        \item Compute cross-entropy loss: $L = - (1/N) \times \sum(y_{\text{true}} * \log(y_{\text{pred}}))$
    \end{enumerate}
    \textbf{Backpropagation and parameter update:}
    \begin{enumerate}
        \item Compute gradients of the loss w.r.t. all parameters
        \item Update parameters using Adam optimizer
    \end{enumerate}
}
\textbf{Prediction:}
$y_{\text{pred}} = \text{argmax}(\text{MPNN}(X, A))$
\end{algorithm}

\begin{figure*}[!ht]
    \centering
    \includegraphics[width=6in]{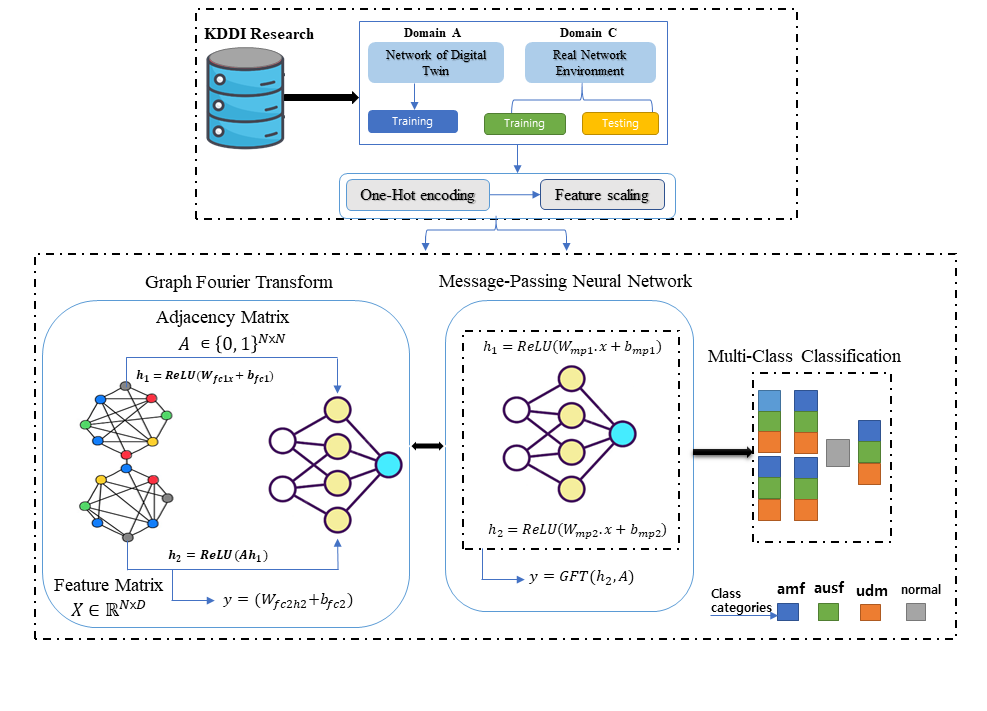}
    \caption{Overall System of the Proposed GFT-MPNN for 5G Core Network}
    \label{fig: B}
\end{figure*}

\section{EXPERIMENTS AND EVALUATION}

\subsection{Datasets and preprocessing}
This section details the data preprocessing stage of the system. This stage consists of two subprocesses: one-hot encoding for categorical classes, and feature scaling for normalization.
The system transforms nominal attributes into one-hot vectors. Each nominal (categorical) attribute is represented as a binary vector with a size equal to the number of attribute values. In this binary vector, only one point corresponds to the expressed value, which is assigned a value of 1, whereas all other points are assigned a value of 0. For instance, for the failure injection attribute commonly used in network failure classification, with the failure points pf the process and mobility management function “amf,” authentication server function “ausf,” and unified data management “udm,” the attribute is transformed into a binary vector of length 3. The attribute values are converted to [1, 0, 0], [0, 1, 0], and [0, 0, 1], respectively.

The system scales the numeric attributes alongside the one-hot encoding process. Normalization methods, such as normalization (e.g., \cite{ieracitano2020novel}) and standardization (e.g., \cite{gao2020omni}) can be considered for scaling numeric features. We adopted the min-max normalization method. The normalization function $\mathcal{f}_A(\cdot)$ for a numeric attribute $A$ that maps every $x$ in $A$ into the range [0, 1] can be defined as follows:
\begin{equation}
f_A(x_i) = \tilde{x}_i = \frac{\text{max}(x_j) - \text{min}(x_j)}{x_i - \text{min}(x_j)},
\end{equation}

where, $x_i$ denotes the $i$th attribute value in attribute $A$.

\begin{itemize}
    \item \textbf{KDDI datasets}: The KDDI dataset is based on the ITU Challenge "ITU-ML5G-PS-008: Network Failure Classification Model Using network digital twin" \cite{Junichi:2023}. The datasets consist of the following: a training dataset (Domain A) derived from NDT simulated environments, a training dataset (Domain C) generated from a real network, and a testing dataset (Domain C) also obtained from the same real network. Each domain dataset comprises 4121 features, with 16 failure classes per failure sample. However, the failure-type classes consist of a "normal" class and 15 other classes, with over 67\% of the data belonging to the "normal" class. In comparison, each of the remaining 15 failure classes comprises only 2.2\% of the datasets. Table \ref{tab: C} represents the distribution table for the three domains.
\end{itemize}

\begin{table}[ht]
    \centering
    \caption{Datasets Distribution Table describing the number of features and the number of samples in the three domains}
    \label{tab: C}
    \begin{tabular}{>{\centering\arraybackslash} p{3cm}|c|c} 
        \toprule
        Datasets & No. of features & No. of samples \\
        \midrule
        Network of Digital Twin Data for Training (Domain A) & 4121 & 80 / failure class \\ \hline
        Real Network Data for Training (Domain C) & 4121 & 10 / failure class \\ \hline
        Real Network Data for Testing (Domain C) & 4121 & 9-10 / failure class \\ \hline
        \bottomrule
    \end{tabular}
\end{table}

    \subsection{Data Imbalance}
    
    \begin{itemize}
     \item \textbf{Handling Data Imbalance:} Data balancing of minority and majority classes is a crucial aspect of data preparation. In comparing the three datasets, including the testing dataset, the number of failure occurrences is much lower, especially in the real network datasets (Domain C). Small imbalances (e.g., 60:40) between the majority and minority classes typically do not affect the learning ability of a model. However, large imbalances (e.g., 90:10) cause the model to struggle to distinguish between classes correctly. In such cases, a classifier with 90\% accuracy often classifies inputs as the majority class \cite{gupta2021data}. Oversampling \cite{das2014racog} or undersampling \cite{arefeen2020neural} techniques can be employed to correct this imbalance. However, the proposed GFT offers a promising approach to addressing data imbalance in multiclass classification tasks.

    \item \textbf{GFT-based Feature Transformation:} To extract meaningful information, we employed the GFT to capture the underlying structural information of the graph data by decomposing the graph features using the eigenvalues and eigenvectors of the Laplacian matrix. The GFT extracts discriminative features that highlight the differences between classes, including both majority and minority classes. This approach enables the proposed model to better understand the data distribution and identify critical patterns. The transformed features provide an enhanced data representation, incorporating local and global structural information to improve classification performance.

    \item \textbf{MPNN-based Feature Extraction:} The MPNN adapts its parameters based on the propagated messages, allowing the model to learn from the graph structure and the features extracted by GFT. The message-passing mechanism facilitates the spreading of information across the graph, including minority class samples. This mechanism ensures that minority classes receive sufficient attention during training, facilitating the learning process. By combining GFT with MPNN, the model learns to extract features that are informative for both the majority and minority classes, mitigating the effects of a data imbalance.
\end{itemize}

This section reviews the target datasets from KDDI research and describes the detailed implementation of each component. Then, the experimental results and comparative analysis are presented, and the proposed systems are evaluated.

\subsection{Experimental Setup}
The GFT is applied as part of the MPNN model architecture. It serves as a feature extraction mechanism that captures specific network dynamics and patterns from the input data. Integrated into the MPNN model architecture as a layer, the GFT is incorporated into the message-passing mechanism of the MPNN. Within the MPNN architecture, the GFT layer extracts features from the input data, representing relevant characteristics of the graph-structured data, such as the relationships between "amf", "ausf", "udm".

The input dimension of the MPNN model is determined by the number of features in the input data and is set to the number of columns (features) in the input datasets. The hidden dimension determines the number of units (neurons) in the hidden layers of the MPNN model, which is set to 64; thus, each hidden layer has 64 units. The output dimension of the MPNN model corresponds to the number of classes or categories in the target variable, which is calculated as the number of unique classes. The performance of the trained MPNN model is evaluated using the accuracy, precision (weighted average), recall (weighted average), and F1-score (weighted average) metrics. Additionally, the training time for feature extraction was observed for Domain A (training) 9.12 s, Domain C (training) 1.45 s, and Domain C (testing) 2.18 s, respectively. Table \ref{tab: Param} represents the training parameters.

\begin{table}[ht]
    \centering
    \caption{Training Parameters}
    \label{tab: Param}
    \begin{tabular}{>{\centering\arraybackslash} p{3cm}|c} 
        \toprule
        Parameters & Value \\
        \midrule
        Dimension & Hidden\_dim=64 \\ \hline
       Learning rate & lr = 0.001  \\ \hline
       Optimizer & Adam optimizer \\ \hline
       Loss function &Cross-entropy loss \\ \hline
       Number of epochs & num\_epochs = 500 \\ \hline
       Dimension & Hidden\_dim \\ \hline
        \bottomrule
    \end{tabular}
\end{table}

\subsection{Performance Metrics}

In the experiment, we measured the performance of the model based on accuracy, precision, recall, and the F1-score to assess their performance. The percentage of correctly inferred outputs is the accuracy, and it is frequently used to measure how well AI models perform. Recall is the proportion of data with positive values that the model properly infers, whereas precision indicates the fraction of positive values inferred by the model that is correct for a given class in a dataset. The harmonic mean of the precision and recall is the F1-score. The following is a define these metrics:

\begin{enumerate}
    \item Accuracy: 
    \begin{equation}
         \text{Accuracy} = \frac{TP + TN}{TP + FP + TN + FN}
    \end{equation}
   
    \item Precision: 
    \begin{equation}
         \text{Precision} = \frac{TP}{TP + FP}
    \end{equation}
   
    \item Recall: 
    \begin{equation}
        \text{Recall} = \frac{TP}{TP + FN}
    \end{equation}
    
    \item F1-score: 
    \begin{equation}
         \text{F1-score} = \frac{2 \times \text{Precision} \times \text{Recall}}{\text{Precision} + \text{Recall}}
    \end{equation}
   
\end{enumerate}

where TP, TN, FN, and FP denote the true positive, true negative, false negative, and false positive values, respectively. In the GFT-MPNN experiments, we evaluated each model using standard performance metrics. The precision, recall, and F1-score metrics vary significantly across classes, indicating differences in how well the model predicts each class. This heterogeneity in performance suggests that the model may perform well for some classes and less effectively for others.

The categorization findings are presented using the receiver operating characteristic (ROC) curve. The TP and FP rates classification performance of a model at an ideal threshold is represented graphically by the ROC. Plotting the TP rate against the FP rate at various thresholds, the ROC curve specifically identifies the ideal threshold \cite{atapattu2010analysis}.


\subsection{Results and Analysis}

\begin{table*}[!ht]
    \centering
    \caption{Classification Results for the training in a Network Digital Twin Environment (Domain A) and Real Network Data (Domain C)}
    \label{tab A}
    \begin{tabular}{c|c|ccc|ccc|ccc}
    
            \hline \hline
            & \multicolumn{1}{c|}{Failure types} & \multicolumn{3}{c}{Domain A (Training)} & \multicolumn{3}{c}{Domain C (Training)} & \multicolumn{3}{c}{Domain C (Testing)} \\ \cline{1-11}
              Classes & Sub-Types & Precision  & Recall & F1-Score & Precision & Recall & F1-Score & Precision & Recall & F1-Score \\
             \hline
0 & amfx1\_bridge-delif & 0.99 & 0.86 & 0.92 & 0.88 & 0.70 & 0.78 & 0.99 & 0.86 & 0.92 \\
1 & amfx1\_ens5\_interface-down & 1.00 & 1.00 & 1.00 & 1.00 & 1.00 & 1.00 & 1.00 & 1.00 & 1.00 \\
2 & amfx1\_ens5\_interface-loss-start-70 & 0.91 & 0.87 & 0.89 & 1.00 & 0.90 & 0.95 & 0.91 & 0.87 & 0.89 \\
3 & amfx1\_memory-stress-start & 1.00 & 0.99 & 0.99 & 1.00 & 1.00 & 1.00 & 1.00 & 0.99 & 0.99 \\

4 & amfx1\_vcpu-overload-start & 0.95 & 0.91 & 0.93 & 1.00 & 0.90 & 0.95 & 0.95 & 0.91 & 0.93 \\
5 & ausfx1\_bridge-delif & 0.68 & 0.52 & 0.59 & 0.82 & 0.90 & 0.86  & 0.68 & 0.52 & 0.59 \\
6 & ausfx1\_ens5\_interface-down & 1.00 & 1.00 &  1.00 & 1.00 & 1.00 & 1.00 & 1.00 & 1.00 &  1.00 \\
7 & ausfx1\_ens5\_interface-loss-start-70 & 0.69 & 0.25 & 0.37 & 1.00 & 1.00 & 1.00 & 0.69 & 0.25 & 0.37 \\
8 & ausfx1\_memory-stress-start & 1.00 & 1.00 & 1.00 & 1.00 & 1.00 & 1.00 & 1.00 & 1.00 & 1.00 \\
9 & ausfx1\_vcpu-overload-start & 1.00 & 1.00 & 1.00 & 1.00 & 1.00 & 1.00 & 1.00 & 1.00 & 1.00 \\
10 & normal & 0.94 & 1.00 & 0.97 & 0.97 & 1.00 & 0.98 & 0.94 & 1.00 & 0.97 \\
11 & udmx1\_bridge-delif & 0.92 & 0.58 & 0.71 & 0.83 & 0.50 & 0.62 & 0.92 & 0.58 & 0.71 \\
12 & udmx1\_ens5\_interface-down & 1.00 & 1.00 & 1.00 & 1.00 & 0.90 & 0.95 & 1.00 & 1.00 & 1.00 \\
13 & udmx1\_ens5\_interface-loss-start-70 & 0.95 & 0.25 & 0.40 & 1.00 & 0.80 & 0.89 & 0.95 & 0.25 & 0.40 \\
14 & udmx1\_memory-stress-start & 1.00 & 1.00 & 1.00 & 1.00 & 1.00 & 1.00 & 1.00 & 1.00 & 1.00 \\
15 & udmx1\_vcpu-overload-start & 0.92 & 1.00 & 0.96 & 1.00 & 1.00 & 1.00 & 0.92 & 1.00 & 0.96 \\
             \hline
      \hline

        & \textbf{Macro avg}  & 0.99 & 0.83 & 0.86 & 0.97 & 0.91 & 0.94 & 0.99 & 0.95 & 0.97 \\
      &  \textbf{weighted avg}  & 0.99 & 0.96 & 0.97 & 0.97 & 0.97 & 0.97 & 0.98 & 0.98 & 0.98 \\
    \hline
    \end{tabular}
\end{table*}

\begin{figure}[!ht]
    \centering
    \includegraphics[width=8.5cm]{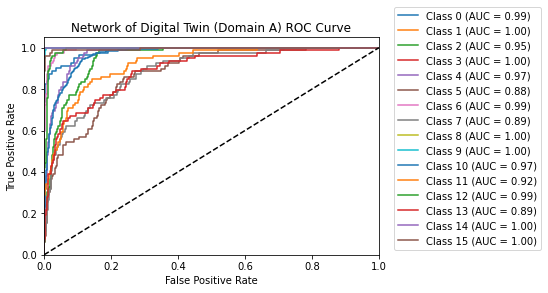}
    \caption{Confusion matrix of the network of digital twin environment (Domain A)}
    \label{fig: Network of Digital Twin (Domain A)}
\end{figure}

The classification reports provide the precision, recall, and the F1-score for every class in the Domain A (training), Domain C (Training), and Domain C (training), and Domain C (testing) datasets, providing a thorough performance analysis across domains. In Table \ref{tab A}, performance on the training data from the NDT environment in Domain A is lower compared to the other two domains in real network environments, with an average F1-score of 0.93. The precision, recall, and accuracy are slightly lower, around 0.94, indicating that the model performs well but may struggle to identify some particular classes. Precision measures the proportion of TP predictions out of all positive predictions (TP and FP) made by the model. In the Domain A dataset, Classes 1, 3, 6, 8, 9, 12, and 14 achieved perfect precision (1.00), indicating that the model correctly identified all instances belonging to these classes among the positive predictions. However, Classes 5, 7, and 13 have relatively lower precision values, suggesting that the model predictions for these classes include more FP.
In the Domain C (training) dataset, as shown in Fig. \ref{fig: Network of Digital Twin (Domain A)}, most classes achieved high precision values, with Classes 1, 2, 3, 4, 6, 7, 8, 9, 10, 12, 13, 14, and 15 achieving perfect precision (1.00). However, Class 0 has a lower precision value, suggesting that the model predictions for this class include more FP.


\begin{figure}[!ht]
    \centering
    \includegraphics[width=8.5cm]{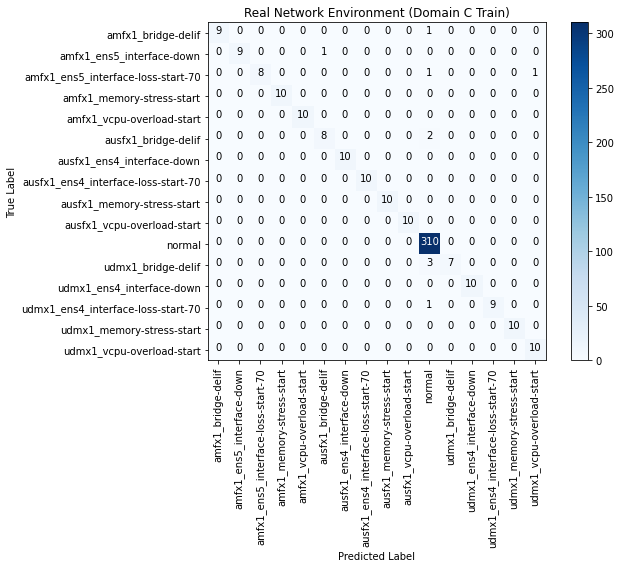}
    \caption{Confusion matrix obtained from the real network (Domain C)}
    \label{fig: Real Network Data (Domain C)}
\end{figure}

\begin{figure}[!ht]
    \centering
    \includegraphics[width=8.5cm]{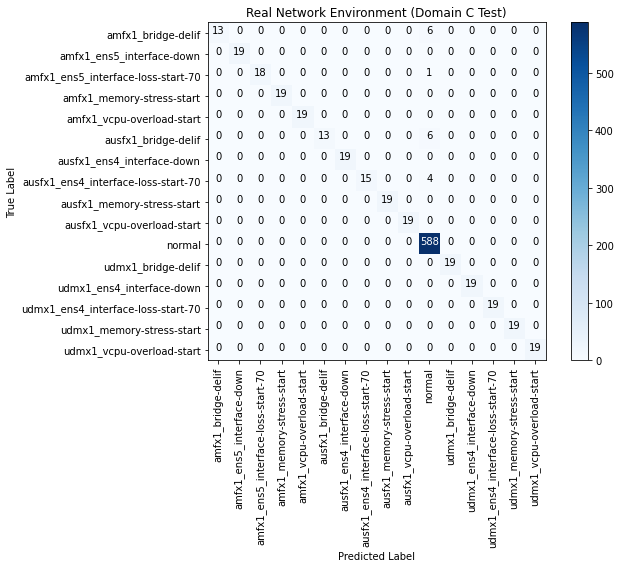}
    \caption{Confusion matrix obtained from the real network testing data (Domain C)}
    \label{fig: Real Network Data Domain C (Testing)}
\end{figure}

Also known as sensitivity, recall measures the proportion of TP predictions out of all actual positive instances (TP and FN) in the dataset. In Fig. 2, classes 1, 3, 6, 8, 9, 12, 14, and 15 achieved perfect recall (1.00), indicating that the model correctly identified all instances of these classes. However, classes 0, 2, 4, 5, 7, 11, and 13 have lower recall values, indicating that the model missed some instances of these classes. In the Domain C (testing) dataset, in Table \ref{tab A}, most classes have high recall values, indicating that the model correctly identified the majority of instances for these classes. However, Class 11 has a lower recall value, suggesting that the model missed some instances for these classes. The F1-score is the harmonic mean of the precision and recall, providing a balance between the two metrics, which is useful for datasets with class imbalance. The weighted average F1-score for the Domain A dataset is 0.93, indicating overall good performance across all classes. The weighted average F1-score for the Domain C (training) dataset is 0.97, indicating overall excellent performance across all classes. Overall, both datasets perform well in terms of precision, recall, and F1-score results for most classes, indicating that the models effectively classify instances across classes. The results suggest that the model accurately identifies and classifies the majority of data points. Class 0 has the lowest recall (0.63) compared to the others, meaning the model might be missing a significant portion of actual Class 0 instances, classifying them into other categories. Similarly, Class 5 also has a lower recall (0.68) and F1-score (0.81). Class 13 has a slightly lower recall (0.95) compared to others. Further investigation into Classes 0, 5, and 13 is recommended to improve the classification of these classes and enhance the overall model effectiveness on the testing dataset.

\begin{figure}
        \centering
        \includegraphics[width=\linewidth]{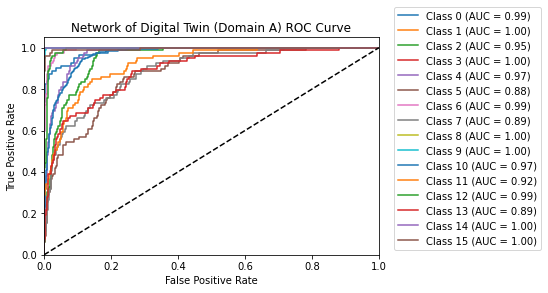}
        \caption{ROC Curve analysis obtained from a network of digital twin}
        \label{fig: ROC Curve analysis obtained from a network of digital twin}
    \end{figure}
    
    \begin{figure}
        \centering
        \includegraphics[width=\linewidth]{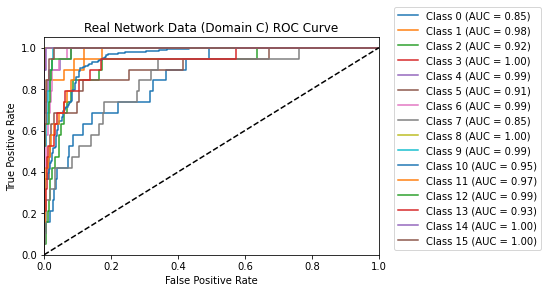}
        \caption{ROC curve analysis obtained from real network (Domain C)}
        \label{fig: ROC Curve analysis obtained from real network dataset}
    \end{figure}

The confusion matrix in Figs. \ref{fig: Network of Digital Twin (Domain A)} and \ref{fig: Real Network Data (Domain C)} indicate the training performance of each failure class in Domains A and Domain C. A high precision indicates that the model makes fewer FP predictions. In Fig. \ref{fig: Network of Digital Twin (Domain A)}, precision varies across classes, with some classes having high precision (e.g., Classes 0, 8, 9, 12, 14, 15) and others having lower precision (e.g., Classes 5, 7, 11, 13). Similarly, in Fig. \ref{fig: Real Network Data (Domain C)}, precision varies across classes, but overall, it is high for most classes.

    \begin{figure}
        \centering
        \includegraphics[width=\linewidth]{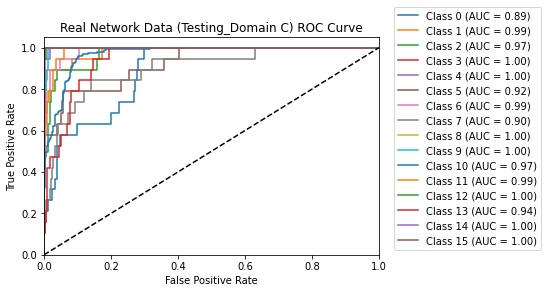}
        \caption{ROC Curve analysis obtained from a real network testing data (Domain C)}
        \label{fig: ROC Curve analysis from the real network testing datasets}
    \end{figure}
    
Fig. \ref{fig: Real Network Data Domain C (Testing)} reveals the confusion matrix analysis of the classification, and the model performs well for most classes but struggles with identifying a portion of the data points that belong to Class 0 (and potentially class 5).

\begin{figure}
    \centering
    \includegraphics[width=8.5cm]{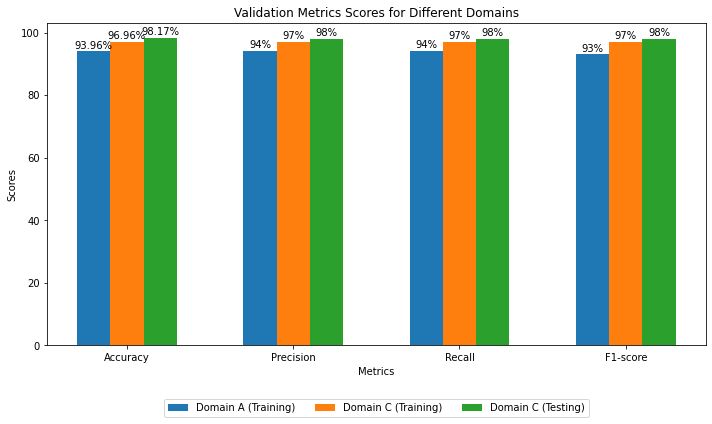}
    \caption{Validation metrics of three datasets}
    \label{fig: Average validation metrics of the three domains}
\end{figure}

The results in Fig. \ref{fig: Average validation metrics of the three domains} indicate the validation performance of a model trained on three domains. In Domain C (testing), the model achieved an accuracy of 98.05\%, indicating that it correctly classified 98.05\% of the samples in the testing dataset. The precision, recall, and F1-score are also high, at 0.98, indicating a high performance across all classes.

In Domain C (training) the performance is slightly lower than on the testing data, with an accuracy of 97.83\%. However, the precision, recall, and F1-score remain consistent at 97\%, indicating robust performance on the training dataset from the real network.

In Domain A (training) the performance is lower than in the other two domains, with an accuracy of 94.27\%. The precision, recall, and F1-score are also slightly lower with an average of 94\%, indicating that the model in the simulated environment (Domain A) struggled to classify some failure types.

\subsection{Discussion}
This work proposes a novel approach combining the GFT MPNN to address the class imbalance problem in multiclass network failure classification. The GFT extracts discriminative features that highlight the differences between the majority and minority classes, enabling the model to better understand the data distribution. Unlike the SMOTETomek method used in \cite{rajak2024fdf}, the integrated GFT-MPNN approach does not rely on oversampling or undersampling techniques. Instead, it employs the strengths of the GFT in capturing the structural information and of the MPNN in modeling dependencies within the network structure. While the work in \cite{liu2021pick} employs a GNN-based approach to address imbalance, our proposed method outperforms it in terms of precision, recall, and F1 score results.
 
The ROC curve analysis, confirms the model convergence across all failure categories, indicating its ability to learn detailed representations from the datasets. However, the confusion matrices derived from the classification and the ROC analysis (Figs \ref{fig: ROC Curve analysis obtained from a network of digital twin}, \ref{fig: ROC Curve analysis obtained from real network dataset}, and \ref{fig: ROC Curve analysis from the real network testing datasets}), suggest potential areas for improvement, particularly in addressing imbalanced data distributions and specific failure classes with lower recall values, such as Class 0 and 5.

The average validation metrics (Fig \ref{fig: Average validation metrics of the three domains}), demonstrate the superior performance of the model in recognizing failure types encountered in real network environments compared to other approaches. This investigation was conducted using the Python programming language in the Spyder IDE environment, without utilizing GPU acceleration during the experimentation process.

To the best of the authors' knowledge, the integration of the GFT into the MPNN for multiclassification of 5G core network failures represents a novel finding, despite the challenges associated with incomplete training data. The proposed model exhibits promising results and highlights the importance of advanced techniques, such as GFT-based architectures, in network failure detection tasks. Continued research and refinement of the model could further enhance its performance and applicability in real-world scenarios.

\section{Conclusion}

This article proposes a novel GFT-MPNN model for multiclass network failure classification in an NDT. The model demonstrates promising potential in addressing the data imbalance problem which is common in NDTs. The GFT-MPNN efficiently extracts meaningful representations (embeddings) from the unbalanced data using the GFT to address imbalances. This approach enables the model to classify network failure instances with high precision and recall across most classes. These results suggest that the GFT plays a crucial role in mitigating the effects of a data imbalance. 
Moreover, by capturing the underlying structure of the network data, the GFT facilitates accurate classification even in skewed class distributions. The proposed model achieved impressive performance on the KDDI datasets, with the validation metrics consistently exceeding an F1-score of 93\% across all domains. The model evaluation on real-world datasets confirms its ability to pinpoint failure types and locations accurately within complex network infrastructures, such as 5G core networks. 

For future work, besides extraction information, applying explainable AI techniques could facilitate understanding how the model makes classifications. This understanding could be valuable for network operators who must interpret and trust the model predictions for real-world decision-making.

\bibliographystyle{IEEEtran} 
\bibliography{Journal_Papers/main} 

\begin{thebibliography}{10}
\providecommand{\url}[1]{#1}
\csname url@samestyle\endcsname
\providecommand{\newblock}{\relax}
\providecommand{\bibinfo}[2]{#2}
\providecommand{\BIBentrySTDinterwordspacing}{\spaceskip=0pt\relax}
\providecommand{\BIBentryALTinterwordstretchfactor}{4}
\providecommand{\BIBentryALTinterwordspacing}{\spaceskip=\fontdimen2\font plus
\BIBentryALTinterwordstretchfactor\fontdimen3\font minus \fontdimen4\font\relax}
\providecommand{\BIBforeignlanguage}[2]{{%
\expandafter\ifx\csname l@#1\endcsname\relax
\typeout{** WARNING: IEEEtran.bst: No hyphenation pattern has been}%
\typeout{** loaded for the language `#1'. Using the pattern for}%
\typeout{** the default language instead.}%
\else
\language=\csname l@#1\endcsname
\fi
#2}}
\providecommand{\BIBdecl}{\relax}
\BIBdecl

\bibitem{Cisco:2020}
\BIBentryALTinterwordspacing
Cisco, ``Cisco annual internet report (2018–2023) white paper,'' Mar. 2020. [Online]. Available: \url{https://www.cisco.com/c/en/us/solutions/collateral/executive-perspectives/annual-internet-report/white-paper-c11-741490.html}
\BIBentrySTDinterwordspacing

\bibitem{wang2022spatial}
Z.~Wang, J.~Hu, G.~Min, Z.~Zhao, Z.~Chang, and Z.~Wang, ``Spatial-temporal cellular traffic prediction for 5g and beyond a graph neural networks-based approach,'' \emph{IEEE Transactions on Industrial Informatics}, vol.~19, no.~4, pp. 5722--5731, 2022.

\bibitem{grieves2014digital}
M.~Grieves, ``Digital twin: manufacturing excellence through virtual factory replication,'' \emph{White paper}, vol.~1, no. 2014, pp. 1--7, 2014.

\bibitem{isah2023digital}
A.~Isah, H.~Shin, S.~Oh, S.~Oh, I.~Aliyu, T.-w. Um, and J.~Kim, ``Digital twins temporal dependencies-based on time series using multivariate long short-term memory,'' \emph{Electronics}, vol.~12, no.~19, p. 4187, 2023.

\bibitem{yu2023digital}
P.~Yu, J.~Zhang, H.~Fang, W.~Li, L.~Feng, F.~Zhou, P.~Xiao, and S.~Guo, ``Digital twin driven service self-healing with graph neural networks in 6g edge networks,'' \emph{IEEE Journal on Selected Areas in Communications}, 2023.

\bibitem{wu2021digital}
Y.~Wu, K.~Zhang, and Y.~Zhang, ``Digital twin networks: A survey,'' \emph{IEEE Internet of Things Journal}, vol.~8, no.~18, pp. 13\,789--13\,804, 2021.

\bibitem{isah2023towards}
A.~Isah, H.~Shin, S.~A. Hassan, S.~Oh, and J.~Kim, ``Towards temporal dependency identification based on multivariate time series iiot data,'' in \emph{2023 14th International Conference on Information and Communication Technology Convergence (ICTC)}.\hskip 1em plus 0.5em minus 0.4em\relax IEEE, 2023, pp. 368--371.

\bibitem{cases2020key}
I.~R.~U. Cases, ``Key network requirements for network 2030,'' \emph{ITU: Geneva, Switzerland}, 2020.

\bibitem{smilkov2017smoothgrad}
D.~Smilkov, N.~Thorat, B.~Kim, F.~Vi{\'e}gas, and M.~Wattenberg, ``Smoothgrad: removing noise by adding noise,'' \emph{arXiv preprint arXiv:1706.03825}, 2017.

\bibitem{zhou2016learning}
B.~Zhou, A.~Khosla, A.~Lapedriza, A.~Oliva, and A.~Torralba, ``Learning deep features for discriminative localization,'' in \emph{Proceedings of the IEEE conference on computer vision and pattern recognition}, 2016, pp. 2921--2929.

\bibitem{selvaraju2017grad}
R.~R. Selvaraju, M.~Cogswell, A.~Das, R.~Vedantam, D.~Parikh, and D.~Batra, ``Grad-cam: Visual explanations from deep networks via gradient-based localization,'' in \emph{Proceedings of the IEEE international conference on computer vision}, 2017, pp. 618--626.

\bibitem{dabkowski2017real}
P.~Dabkowski and Y.~Gal, ``Real time image saliency for black box classifiers,'' \emph{Advances in neural information processing systems}, vol.~30, 2017.

\bibitem{yuan2020interpreting}
H.~Yuan, L.~Cai, X.~Hu, J.~Wang, and S.~Ji, ``Interpreting image classifiers by generating discrete masks,'' \emph{IEEE Transactions on Pattern Analysis and Machine Intelligence}, vol.~44, no.~4, pp. 2019--2030, 2020.

\bibitem{olah2017feature}
C.~Olah, A.~Mordvintsev, and L.~Schubert, ``Feature visualization,'' \emph{Distill}, vol.~2, no.~11, p.~e7, 2017.

\bibitem{olah2018building}
C.~Olah, A.~Satyanarayan, I.~Johnson, S.~Carter, L.~Schubert, K.~Ye, and A.~Mordvintsev, ``The building blocks of interpretability,'' \emph{Distill}, vol.~3, no.~3, p. e10, 2018.

\bibitem{yang2019xfake}
F.~Yang, S.~K. Pentyala, S.~Mohseni, M.~Du, H.~Yuan, R.~Linder, E.~D. Ragan, S.~Ji, and X.~Hu, ``Xfake: Explainable fake news detector with visualizations,'' in \emph{The world wide web conference}, 2019, pp. 3600--3604.

\bibitem{du2018towards}
M.~Du, N.~Liu, Q.~Song, and X.~Hu, ``Towards explanation of dnn-based prediction with guided feature inversion,'' in \emph{Proceedings of the 24th ACM SIGKDD International Conference on Knowledge Discovery \& Data Mining}, 2018, pp. 1358--1367.

\bibitem{wang2020icapsnets}
Z.~Wang, X.~Hu, and S.~Ji, ``icapsnets: Towards interpretable capsule networks for text classification,'' \emph{arXiv preprint arXiv:2006.00075}, 2020.

\bibitem{du2019attribution}
M.~Du, N.~Liu, F.~Yang, S.~Ji, and X.~Hu, ``On the attribution of recurrent neural network predictions via additive decomposition,'' in \emph{The World Wide Web Conference}, 2019, pp. 383--393.

\bibitem{yuan2019interpreting}
H.~Yuan, Y.~Chen, X.~Hu, and S.~Ji, ``Interpreting deep models for text analysis via optimization and regularization methods,'' in \emph{Proceedings of the AAAI Conference on Artificial Intelligence}, vol.~33, no.~01, 2019, pp. 5717--5724.

\bibitem{du2019techniques}
M.~Du, N.~Liu, and X.~Hu, ``Techniques for interpretable machine learning,'' \emph{Communications of the ACM}, vol.~63, no.~1, pp. 68--77, 2019.

\bibitem{isah2024graph}
A.~Isah, H.~Shin, I.~Aliyu, R.~M. Sulaiman, and J.~Kim, ``Graph neural network for digital twin network: A conceptual framework,'' in \emph{2024 International Conference on Artificial Intelligence in Information and Communication (ICAIIC)}.\hskip 1em plus 0.5em minus 0.4em\relax IEEE, 2024, pp. 1--5.

\bibitem{ferriol2023routenet}
M.~Ferriol-Galm{\'e}s, J.~Paillisse, J.~Su{\'a}rez-Varela, K.~Rusek, S.~Xiao, X.~Shi, X.~Cheng, P.~Barlet-Ros, and A.~Cabellos-Aparicio, ``Routenet-fermi: Network modeling with graph neural networks,'' \emph{IEEE/ACM transactions on networking}, 2023.

\bibitem{galmes2022routenet}
M.~F. Galm{\'e}s, J.~Pailliss{\'e}, J.~Su{\'a}rez-Varela, K.~Rusek, S.~Xiao, X.~Shi, X.~Cheng, P.~Barlet-Ros, and A.~Cabellos-Aparicio, ``Routenet-fermi: Network modeling with graph neural networks.'' \emph{arXiv preprint arXiv:2212.12070}, 2022.

\bibitem{isah2023datadriven}
A.~Isah, H.~Shin, I.~Aliyu, S.~Oh, S.~Lee, J.~Park, M.~Hahn, and J.~Kim, ``A data-driven digital twin network architecture in the industrial internet of things (iiot) applications,'' 2023.

\bibitem{zhang2024knowledge}
X.~Zhang, L.~Zheng, W.~Fan, W.~Ji, L.~Mao, and L.~Wang, ``Knowledge graph and function block based digital twin modeling for robotic machining of large-scale components,'' \emph{Robotics and Computer-Integrated Manufacturing}, vol.~85, p. 102609, 2024.

\bibitem{liu2020alleviating}
Z.~Liu, Y.~Dou, P.~S. Yu, Y.~Deng, and H.~Peng, ``Alleviating the inconsistency problem of applying graph neural network to fraud detection,'' in \emph{Proceedings of the 43rd international ACM SIGIR conference on research and development in information retrieval}, 2020, pp. 1569--1572.

\bibitem{peng2019trainable}
M.~Peng, Q.~Zhang, X.~Xing, T.~Gui, X.~Huang, Y.-G. Jiang, K.~Ding, and Z.~Chen, ``Trainable undersampling for class-imbalance learning,'' in \emph{Proceedings of the AAAI conference on artificial intelligence}, vol.~33, no.~01, 2019, pp. 4707--4714.

\bibitem{huang2022graph}
Z.~Huang, Y.~Tang, and Y.~Chen, ``A graph neural network-based node classification model on class-imbalanced graph data,'' \emph{Knowledge-Based Systems}, vol. 244, p. 108538, 2022.

\bibitem{habibi2019comprehensive}
M.~A. Habibi, M.~Nasimi, B.~Han, and H.~D. Schotten, ``A comprehensive survey of ran architectures toward 5g mobile communication system,'' \emph{Ieee Access}, vol.~7, pp. 70\,371--70\,421, 2019.

\bibitem{islam2022deep}
M.~A. Islam, H.~Siddique, W.~Zhang, and I.~Haque, ``A deep neural network-based communication failure prediction scheme in 5g ran,'' \emph{IEEE Transactions on Network and Service Management}, 2022.

\bibitem{said2020network}
M.~Said~Elsayed, N.-A. Le-Khac, S.~Dev, and A.~D. Jurcut, ``Network anomaly detection using lstm based autoencoder,'' in \emph{Proceedings of the 16th ACM Symposium on QoS and Security for Wireless and Mobile Networks}, 2020, pp. 37--45.

\bibitem{shojaee2020safeguard}
M.~Shojaee, M.~Neves, and I.~Haque, ``Safeguard: Congestion and memory-aware failure recovery in sd-wan,'' in \emph{2020 16th International Conference on Network and Service Management (CNSM)}.\hskip 1em plus 0.5em minus 0.4em\relax IEEE, 2020, pp. 1--7.

\bibitem{soltanzadeh2021rcsmote}
P.~Soltanzadeh and M.~Hashemzadeh, ``Rcsmote: Range-controlled synthetic minority over-sampling technique for handling the class imbalance problem,'' \emph{Information Sciences}, vol. 542, pp. 92--111, 2021.

\bibitem{ding2018opportunities}
A.~Y. Ding and M.~Janssen, ``Opportunities for applications using 5g networks: Requirements, challenges, and outlook,'' in \emph{Proceedings of the Seventh International Conference on Telecommunications and Remote Sensing}, 2018, pp. 27--34.

\bibitem{bruna2013spectral}
J.~Bruna, W.~Zaremba, A.~Szlam, and Y.~LeCun, ``Spectral networks and locally connected networks on graphs,'' \emph{arXiv preprint arXiv:1312.6203}, 2013.

\bibitem{defferrard2016convolutional}
M.~Defferrard, X.~Bresson, and P.~Vandergheynst, ``Convolutional neural networks on graphs with fast localized spectral filtering,'' \emph{Advances in neural information processing systems}, vol.~29, 2016.

\bibitem{scarselli2008graph}
F.~Scarselli, M.~Gori, A.~C. Tsoi, M.~Hagenbuchner, and G.~Monfardini, ``The graph neural network model,'' \emph{IEEE Transactions on neural networks}, vol.~20, no.~1, pp. 61--80, 2008.

\bibitem{lei2017deriving}
T.~Lei, W.~Jin, R.~Barzilay, and T.~Jaakkola, ``Deriving neural architectures from sequence and graph kernels,'' in \emph{International Conference on Machine Learning}.\hskip 1em plus 0.5em minus 0.4em\relax PMLR, 2017, pp. 2024--2033.

\bibitem{li2015gated}
Y.~Li, D.~Tarlow, M.~Brockschmidt, and R.~Zemel, ``Gated graph sequence neural networks,'' \emph{arXiv preprint arXiv:1511.05493}, 2015.

\bibitem{kipf2016semi}
T.~N. Kipf and M.~Welling, ``Semi-supervised classification with graph convolutional networks,'' \emph{arXiv preprint arXiv:1609.02907}, 2016.

\bibitem{duvenaud2015convolutional}
D.~K. Duvenaud, D.~Maclaurin, J.~Iparraguirre, R.~Bombarell, T.~Hirzel, A.~Aspuru-Guzik, and R.~P. Adams, ``Convolutional networks on graphs for learning molecular fingerprints,'' \emph{Advances in neural information processing systems}, vol.~28, 2015.

\bibitem{atwood2016diffusion}
J.~Atwood and D.~Towsley, ``Diffusion-convolutional neural networks,'' \emph{Advances in neural information processing systems}, vol.~29, 2016.

\bibitem{niepert2016learning}
M.~Niepert, M.~Ahmed, and K.~Kutzkov, ``Learning convolutional neural networks for graphs,'' in \emph{International conference on machine learning}.\hskip 1em plus 0.5em minus 0.4em\relax PMLR, 2016, pp. 2014--2023.

\bibitem{li2021novel}
J.~Li, Q.~Zhu, Q.~Wu, and Z.~Fan, ``A novel oversampling technique for class-imbalanced learning based on smote and natural neighbors,'' \emph{Information Sciences}, vol. 565, pp. 438--455, 2021.

\bibitem{johnson2019survey}
J.~M. Johnson and T.~M. Khoshgoftaar, ``Survey on deep learning with class imbalance,'' \emph{Journal of Big Data}, vol.~6, no.~1, pp. 1--54, 2019.

\bibitem{chawla2002smote}
N.~V. Chawla, K.~W. Bowyer, L.~O. Hall, and W.~P. Kegelmeyer, ``Smote: synthetic minority over-sampling technique,'' \emph{Journal of artificial intelligence research}, vol.~16, pp. 321--357, 2002.

\bibitem{koziarski2019radial}
M.~Koziarski, B.~Krawczyk, and M.~Wo{\'z}niak, ``Radial-based oversampling for noisy imbalanced data classification,'' \emph{Neurocomputing}, vol. 343, pp. 19--33, 2019.

\bibitem{han2005borderline}
H.~Han, W.-Y. Wang, and B.-H. Mao, ``Borderline-smote: a new over-sampling method in imbalanced data sets learning,'' in \emph{International conference on intelligent computing}.\hskip 1em plus 0.5em minus 0.4em\relax Springer, 2005, pp. 878--887.

\bibitem{bunkhumpornpat2009safe}
C.~Bunkhumpornpat, K.~Sinapiromsaran, and C.~Lursinsap, ``Safe-level-smote: Safe-level-synthetic minority over-sampling technique for handling the class imbalanced problem,'' in \emph{Advances in Knowledge Discovery and Data Mining: 13th Pacific-Asia Conference, PAKDD 2009 Bangkok, Thailand, April 27-30, 2009 Proceedings 13}.\hskip 1em plus 0.5em minus 0.4em\relax Springer, 2009, pp. 475--482.

\bibitem{gilmer2017neural}
J.~Gilmer, S.~S. Schoenholz, P.~F. Riley, O.~Vinyals, and G.~E. Dahl, ``Neural message passing for quantum chemistry,'' in \emph{International conference on machine learning}.\hskip 1em plus 0.5em minus 0.4em\relax PMLR, 2017, pp. 1263--1272.

\bibitem{lin2017focal}
T.-Y. Lin, P.~Goyal, R.~Girshick, K.~He, and P.~Doll{\'a}r, ``Focal loss for dense object detection,'' in \emph{Proceedings of the IEEE international conference on computer vision}, 2017, pp. 2980--2988.

\bibitem{wang2016training}
S.~Wang, W.~Liu, J.~Wu, L.~Cao, Q.~Meng, and P.~J. Kennedy, ``Training deep neural networks on imbalanced data sets,'' in \emph{2016 international joint conference on neural networks (IJCNN)}.\hskip 1em plus 0.5em minus 0.4em\relax IEEE, 2016, pp. 4368--4374.

\bibitem{shi2020multi}
M.~Shi, Y.~Tang, X.~Zhu, D.~Wilson, and J.~Liu, ``Multi-class imbalanced graph convolutional network learning,'' in \emph{Proceedings of the Twenty-Ninth International Joint Conference on Artificial Intelligence (IJCAI-20)}, 2020.

\bibitem{zhao2021graphsmote}
T.~Zhao, X.~Zhang, and S.~Wang, ``Graphsmote: Imbalanced node classification on graphs with graph neural networks,'' in \emph{Proceedings of the 14th ACM international conference on web search and data mining}, 2021, pp. 833--841.

\bibitem{tan2015online}
M.~Tan, L.~Tan, S.~Dara, and C.~Mayeux, ``Online defect prediction for imbalanced data,'' in \emph{2015 IEEE/ACM 37th IEEE International Conference on Software Engineering}, vol.~2.\hskip 1em plus 0.5em minus 0.4em\relax IEEE, 2015, pp. 99--108.

\bibitem{Albert:2023}
C.~Albert~Cabellos, ``2023 graph neural networking challenge | building a network digital twin using data from real networks,'' \url{https://www.youtube.com/watch?v=QzDTYyWzpr0&t=2792s}, Jul. 2023.

\bibitem{Junichi:2023}
J.~Kawasaki and N.~Fukumoto, ``2023 japan challenge | network failure classification using network digital twin,'' \url{https://www.youtube.com/watch?v=S4gBXRHQx1o}, Sep. 2023.

\bibitem{lessan2019hybrid}
J.~Lessan, L.~Fu, and C.~Wen, ``A hybrid bayesian network model for predicting delays in train operations,'' \emph{Computers \& Industrial Engineering}, vol. 127, pp. 1214--1222, 2019.

\bibitem{kurokawa2017multi}
T.~Kurokawa, T.~Oki, and H.~Nagao, ``Multi-dimensional graph fourier transform,'' \emph{arXiv preprint arXiv:1712.07811}, 2017.

\bibitem{ieracitano2020novel}
C.~Ieracitano, A.~Adeel, F.~C. Morabito, and A.~Hussain, ``A novel statistical analysis and autoencoder driven intelligent intrusion detection approach,'' \emph{Neurocomputing}, vol. 387, pp. 51--62, 2020.

\bibitem{gao2020omni}
J.~Gao, L.~Gan, F.~Buschendorf, L.~Zhang, H.~Liu, P.~Li, X.~Dong, and T.~Lu, ``Omni scada intrusion detection using deep learning algorithms,'' \emph{IEEE Internet of Things Journal}, vol.~8, no.~2, pp. 951--961, 2020.

\bibitem{gupta2021data}
N.~Gupta, S.~Mujumdar, H.~Patel, S.~Masuda, N.~Panwar, S.~Bandyopadhyay, S.~Mehta, S.~Guttula, S.~Afzal, R.~Sharma~Mittal \emph{et~al.}, ``Data quality for machine learning tasks,'' in \emph{Proceedings of the 27th ACM SIGKDD conference on knowledge discovery \& data mining}, 2021, pp. 4040--4041.

\bibitem{das2014racog}
B.~Das, N.~C. Krishnan, and D.~J. Cook, ``Racog and wracog: Two probabilistic oversampling techniques,'' \emph{IEEE transactions on knowledge and data engineering}, vol.~27, no.~1, pp. 222--234, 2014.

\bibitem{arefeen2020neural}
M.~A. Arefeen, S.~T. Nimi, and M.~S. Rahman, ``Neural network-based undersampling techniques,'' \emph{IEEE Transactions on Systems, Man, and Cybernetics: Systems}, vol.~52, no.~2, pp. 1111--1120, 2020.

\bibitem{atapattu2010analysis}
S.~Atapattu, C.~Tellambura, and H.~Jiang, ``Analysis of area under the roc curve of energy detection,'' \emph{IEEE Transactions on wireless communications}, vol.~9, no.~3, pp. 1216--1225, 2010.

\bibitem{rajak2024fdf}
A.~Rajak and R.~Tripathi, ``Fdf-hybridfs: Towards design of a failure detection framework using hybrid feature selection method for ip core networks that connect 5g core in nfv-based test environment,'' \emph{Computer Standards \& Interfaces}, vol.~87, p. 103779, 2024.

\bibitem{liu2021pick}
Y.~Liu, X.~Ao, Z.~Qin, J.~Chi, J.~Feng, H.~Yang, and Q.~He, ``Pick and choose: a gnn-based imbalanced learning approach for fraud detection,'' in \emph{Proceedings of the web conference 2021}, 2021, pp. 3168--3177.

\end{thebibliography}

\begin{IEEEbiography}
[{\includegraphics[width=1in,height=1.25in,clip,keepaspectratio]{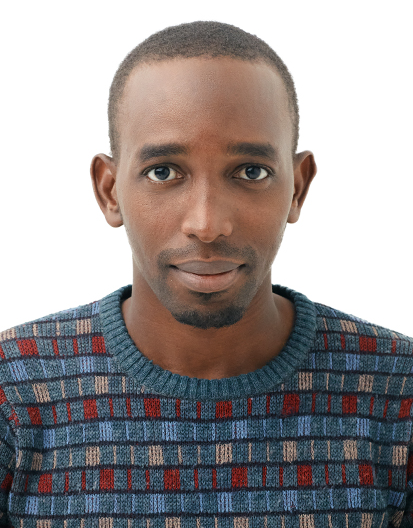}}]
 {Abubakar Isah} received a B.S. degree from Kano University of Science and Technology, Wudil Nigeria, in 2012 and an M.S. degree in computer science from Jodhpur National University, Rajasthan, India. He is currently pursuing a Ph.D. degree with the Hyper Intelligence Network Media Platform (Hi-IoP) Lab, at the Department of Intelligent Electronics and Computer Engineering, Chonnam National University, Gwangju, South Korea. His research interests include AI for autonomous networks, IoT, and data-driven Digital Twin Networks.
\end{IEEEbiography}

\begin{IEEEbiography}
[{\includegraphics[width=1in,height=1.25in,clip,keepaspectratio]{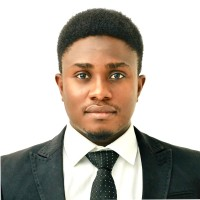}}]
{Ibrahim Aliyu}(Member, IEEE) received the B.Eng. and M.Eng. degrees in computer engineering from the Federal University of Technology, Minna, Nigeria, in 2014 and 2018, respectively, and a Ph.D. degree in computer science and engineering from Chonnam National University, South Korea, in 2022. He is currently a postdoctoral researcher at the Hyper Intelligence Media Network Platform Laboratory, Department of Intelligent Electronics and Computer System Engineering, Chonnam National University, Gwangju, South Korea. His research interests include source routing, in-network computing, and cloud-based computing for massive metaverse deployment. His other research interests include federated learning, data privacy, network security, and AI for autonomous networks. He was a recipient of the 2017 Korean Government Scholarship
Program Award.
\end{IEEEbiography}

\begin{IEEEbiography}
[{\includegraphics[width=1in,height=1.25in,clip,keepaspectratio]{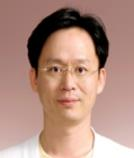}}]
{Jaechan Shim} received the B.S. and M.S. degrees in computer science from the Chungnam National University, Chungnam, South Korea, in 1992 and 1994, respectively, and the Ph.D. degree in computer and engineering from Chonnam National University, South Korea, in 2016. He is currently a Senior Researcher at the Electronics and Telecommunications Research Institute, in South Korea. His research interests are in the Network area. His other research interests include Cloud Computing, High Availability systems, and Digital Twin Networks.
\end{IEEEbiography}

\begin{IEEEbiography}
[{\includegraphics[width=1in,height=1.25in,clip,keepaspectratio]{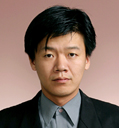}}]
{Hoyong Ryu} received the B.S. and M.S. degrees in Electronics and Communications Engineering from the Kwangwoon University, Seoul, South Korea, in 1991 and 1993, respectively, and the Ph.D. degree in Electronics and Communications Engineering from Kwangwoon University, Seoul, South Korea, in 1999. He is currently a Senior Researcher at the Electronics and Telecommunications Research Institute, in South Korea. His research interests are in the Network area. His other research interests include Cloud Computing, High Availability systems, and Digital Twin Networks.
\end{IEEEbiography}

\begin{IEEEbiography}
[{\includegraphics[width=1in,height=1.25in,clip,keepaspectratio]{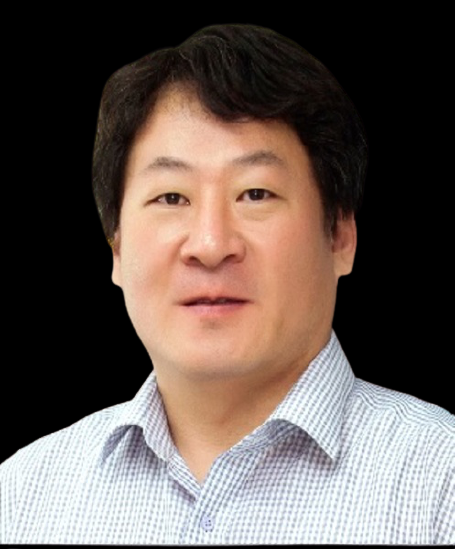}}]
{Jinsul Kim} (Member, IEEE) received a B.S. degree in computer science from the University of Utah, Salt Lake City, Utah, USA, in 1998, and M.S. and Ph.D. degrees in digital media engineering from the Department of Information and Communications, Korea Advanced Institute of Science and Technology (KAIST), Daejeon, South Korea, in 2004 and 2008, respectively. He worked as a researcher at the IPTV Infrastructure Technology Research Laboratory, Broadcasting/Telecommunications Convergence Research Division, Electronics and Telecommunications Research Institute (ETRI), Daejeon, from 2004 to 2009. He also worked as a professor at Korea Nazarene University, Cheonan, South Korea, from 2009 to 2011. He is currently a professor at Chonnam National University, Gwangju, South Korea. His research interests include QoS/QoE, measurement/management, mobile IPTV, social TV, cloud computing, multimedia communication, and digital media arts. He has been an invited reviewer of IEEE Transactions on Multimedia, since 2008. He has been invited to the Technical Program Committee for IWITMA2009/2010, as the program chair for ICCCT2011 and IWMWT2013/2014/2015, and as the general chair for ICMWT2014.
\end{IEEEbiography}

\vfill

\end{document}